\documentclass[sigconf]{aamas}

\usepackage{amssymb}
\usepackage{graphicx}
\usepackage{algorithm}
\usepackage[noend]{algpseudocode}

\renewcommand\footnotetextcopyrightpermission[1]{} 

\begin{document}

\title{Reinforcement Learning via Reasoning from Demonstration}
\author{Lisa Torrey\\St. Lawrence University}

\begin{abstract}
Demonstration is an appealing way for humans to provide assistance to reinforcement-learning agents. Most approaches in this area view demonstrations primarily as sources of behavioral bias. But in sparse-reward tasks, humans seem to treat demonstrations more as sources of causal knowledge. This paper proposes a framework for agents that benefit from demonstration in this human-inspired way. In this framework, agents develop causal models through observation, and reason from this knowledge to decompose tasks for effective reinforcement learning. Experimental results show that a basic implementation of Reasoning from Demonstration (RfD) is effective in a range of sparse-reward tasks.
\end{abstract}

\keywords{}

\maketitle

\section{Introduction}

With reinforcement learning (RL), agents can train themselves autonomously to complete tasks. RL algorithms can allow agents to develop successful behaviors based only on environmental feedback. However, tasks in large environments with infrequent feedback, known as sparse-reward tasks, have always been difficult for RL agents to learn.

Demonstration is a practical and human-motivated approach to coping with challenging RL tasks. Learning from demonstration (LfD) can either replace RL or augment it. Systems that combine LfD and RL typically do so by encouraging agents to behave more like demonstrators as they learn. This usually improves their early performance, which would otherwise be essentially random.

It can be effective to use demonstrations as models of behavior. This paper asks whether it could also be effective to use demonstrations as sources of knowledge. In sparse-reward tasks, knowledge seems particularly valuable.

Consider, in the classic Atari game Montezuma's Revenge (see Figure \ref{tasks}), the task of exiting the first room without losing a life. It takes a sequence of several hundred carefully chosen actions to complete this task. The deep Q-learning system that initially mastered many Atari games \cite{Mnih2015} did not learn to complete it. Yet the task is simple for a human to understand: the player must navigate to a key, and then to a door, without falling off any platforms, and without colliding with a rolling skull.

This paper proposes a framework in which agents acquire this kind of high-level knowledge by observing a demonstration, and then to use that knowledge to learn the task more effectively.

Recent studies show that rendering video games with textured cells rather than recognizable objects hinders human learning significantly \cite{Dubey2018}. One plausible explanation is that object perception facilitates causal model-building, which cognitive scientists recognize as a critical component of human learning \cite{Lake2017}. This paper asks how LfD agents might incorporate causal reasoning, and thereby benefit from demonstration more like humans seem to do in tasks like Montezuma's Revenge.

\section{Background}

This section presents a short introduction to RL and surveys techniques for making it more effective in challenging tasks. It focuses on single agents learning single tasks, and even within this limited context it is far from exhaustive, but it provides the necessary context for the proposals that follow.

\subsection{Reinforcement learning}

An RL agent develops a policy for choosing actions based on the state of its environment. It receives a numeric reward from the environment as feedback for each step. Balancing between exploiting its developing policy and exploring alternatives, the agent gradually improves its policy and becomes capable of earning higher cumulative rewards \cite{Sutton2018}.

Q-learning \cite{Watkins1992} is one of the simpler RL algorithms. It uses a function $Q(s,a)$ to estimate the cumulative reward an agent can expect to earn after taking action $a$ from state $s$. Given an accurate Q-function, the optimal action in state $s$ is the one that maximizes $Q(s,a)$. Agents can begin with all $Q(s,a) \approx 0$ and then adjust the Q-values incrementally. After taking action $a$ in state $s$, receiving reward $r$, and transitioning to state $s'$, the Q-learning update is:

\begin{equation}
Q(s, a) \longleftarrow (1 - \alpha) Q(s, a) + \alpha (r + \gamma \textrm{max}_{a'} Q(s', a'))
\label{Q1}
\end{equation}

The $\alpha$ parameter is a learning rate, which controls the update size. The $\gamma$ parameter is a discount factor, which controls the agent's patience for delayed rewards. A simple exploration strategy for Q-learning is $\epsilon$-greedy action selection: the agent usually takes the action with the highest Q-value, but with probability $\epsilon$ it chooses a random action instead.

\subsection{Sparse rewards}

RL excels in environments with frequent informative rewards, which allow agents to improve their policies steadily. As non-zero rewards become less frequent, it takes longer to discover them, and it takes more repetition to determine which actions are responsible for them. Sparse rewards cause combinatorial explosions in the training time required to develop competent agents.

\begin{figure*}
\includegraphics[width=\textwidth]{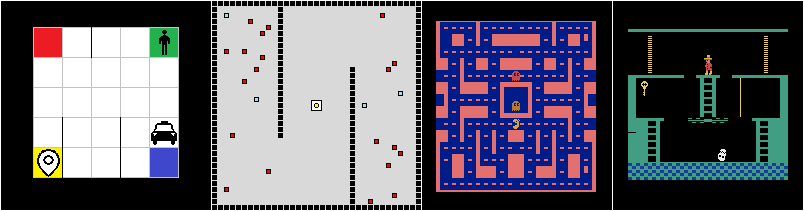}
\caption{Renderings of four virtual environments: Taxi, Courier, Ms. Pacman, and Montezuma's Revenge.}
\label{tasks}
\end{figure*}

Taxi \cite{Dietterich2000} is a small example of a sparse-reward task. In this task, a taxi moves through a 5x5 grid world (see Figure \ref{tasks}), picking up a passenger at one stop and dropping it off at another, with both stops randomly selected from the four corners of the grid. The agent does receive a small negative reward (-1) after each action, to discourage loitering, but it only receives a positive reward (+20) once it completes the task. Each episode ends after 200 actions, whether or not the task is complete.

Courier (see Figure \ref{tasks}) is a scaled-up version of Taxi, with similar dynamics but significantly increased size and sparsity. In this task, a courier moves through a 35x35 grid world, collecting four randomly-placed packages and delivering them to a central platform. There is no time limit on episodes, but there are 20 randomly-placed moving vehicles that the agent must dodge. The only feedback occurs at the end of the task: \emph{success} when the courier delivers the last package, or \emph{failure} if it gets run over.

More examples of sparse-reward tasks can be found in the Arcade Learning Environment \cite{Machado2018}, which emulates classic Atari games with large and unstructured state spaces (160x210 pixel arrays). Figure \ref{tasks} illustrates two of these games. Montezuma's Revenge has one of the most naturally sparse reward systems; in the first room, the only rewards are +100 for acquiring the key and +300 for exiting the room. Ms. Pacman has frequent small rewards for eating pellets, but the largest rewards can only be achieved by generating and eating multiple edible ghosts.

\subsection{Scaling techniques}

Standard Q-learning, which assigns a value to every state-action pair, is only practical in small environments like Taxi. Larger environments, like Courier and the Atari games, have too many states. This problem can be addressed by decomposing states into feature vectors and replacing Q-tables with approximators, which require less memory and allow for generalization across states. Q-learning adapts well to linear and neural approximators using updates derived from gradient descent \cite{Sutton2018}.

There is a long history in RL of engineering features that alter the state space, either to abstract away irrelevant information or to compose useful information. For example, with sophisticated features, linear Q-learning is sufficient to achieve high scores in the first level of Ms. Pacman \cite{Torrey2013}. However, in the era of deep learning, the field has moved away from human feature engineering in favor of implicit feature construction via convolutional neural networks. Deep Q-networks can reach human-level play in many Atari games directly from their raw pixel states \cite{Mnih2015}. But games like Montezuma's Revenge remain difficult for deep Q-learning because of their sparse rewards.

One way to cope with sparse rewards is to shape them. Reward shaping is a technique for providing extra rewards to encourage progress towards task completion. Shaped rewards must be designed appropriately in order to be effective \cite{Ng1999}. For example, on an open grid, it would be effective to reward a taxi by the amount that it reduces its Manhattan distance from its destination, but this intuitive shaping strategy would be counter-productive on the actual Taxi grid, which contains walls that restrict movement. A recent and more abstract form of reward shaping, called intrinsic motivation, encourages exploration by rewarding agents for reaching new areas of the state space; this technique has some success in Montezuma's Revenge \cite{Dann2019}.

Another way to cope with sparsity is to decompose tasks into subtasks and apply hierarchical learning. In the options framework \cite{Sutton1999}, subtask policies are treated like extended actions, which may sometimes be selected instead of primitive actions. Other frameworks, like Feudal RL \cite{Dayan1992} and MAXQ \cite{Dietterich2000}, more explicitly formulate tasks as hierarchies of subtasks.

Like features and rewards, task decompositions have often been engineered. For example, one approach uses human domain knowledge to define a separate learner for each reward source, and chooses actions by aggregating their Q-values \cite{vanSeijen2017}. Given some human engineering of states and rewards as well, this system is capable of outplaying humans in Ms. Pacman. Other approaches train low-level learners to control transitions in a higher-level plan, which is derived from abstract states \cite{Roderick2018} or symbolic rules \cite{Lyu2019} defined by a human. Both of these techniques are effective in Montezuma's Revenge.

There are also approaches in which agents learn to decompose tasks themselves. For example, one method for learning options has been applied to Ms. Pacman \cite{Bacon2017}, and one for learning feudal hierarchies has been applied to Montezuma's Revenge \cite{Vezhnevets2017}. These systems can outperform their non-hierchical baselines, but their scores do not yet compete with the human-guided approaches.

\subsection{Objects}

Many tasks are defined around objects. Important objects tend to be visually displayed for human players, as in Figure \ref{tasks}, but they are not always explicitly described to RL agents. For example, Taxi traditionally decomposes states into four  features: the row and column numbers of the taxi location, and the stop numbers of the passenger and destination. These values uniquely specify each state, but leave much of the visual information implicit. At the other extreme, the pixel states of Atari games provide complete visual information, but in a completely unstructured form.

Some RL algorithms are designed specifically for object-oriented states. Relational RL \cite{Dzeroski2001} trains a first-order logical Q-function approximator, using states that are sets of first-order predicates, grounded with objects in the environment. Object-oriented RL \cite{Diuk2008} uses a similar state representation, though not a first-order Q-function, to speed up learning in Taxi.

Object-oriented states can provide natural opportunities for task decomposition. For example, Object-Focused Q-learning \cite{Cobo2013} trains a policy for interacting with each type of object, and chooses actions by aggregating their Q-values. Hierarchical Deep Q-learning \cite{Kulkarni2016} trains one network to select an object, and another to navigate towards a selected object; this is another approach that is effective in Montezuma's Revenge.

Of all the forms of state engineering involved in RL, object-oriented states are arguably the most intuitive, since humans perceive objects with so little apparent effort. Furthermore, object tracking is an active area of research in the field of  computer vision, and there have been steps towards applying it to object-focused Q-learning \cite{Garnelo2016}.

\subsection{Demonstration}

LfD originated in the field of robotics, where demonstrations are a natural form of assistance for humans to provide, and it can be particularly slow or costly to allow agents to explore autonomously. In this context, LfD may be a form of supervised learning, with demonstrations as the source of training data. For example, an agent can approximate the demonstrator's policy, and use the approximation as its own policy \cite{Sammut1992}.

In environments where exploration is permissible, LfD may also involve autonomous practice. RL provides opportunities to acquire better and more complete policies than can be acquired from demonstration alone. For example, an agent can use an approximate demonstrator policy as a starting point, and then train it further with RL \cite{Schaal1997}. Alternatively, the agent can develop its own policy from scratch, but use an approximate demonstrator policy to choose some actions in the early stages of training \cite{Taylor2011}. Failed demonstrations as well as successful ones can be leveraged to form useful training biases \cite{Grollman2012}.

Demonstrations can also bias the policies of RL agents less directly. For example, one approach shapes rewards for agent actions based on how similar they are to demonstrated actions \cite{Brys2015}. Another system includes demonstrated transitions in the batches that it collects to update a deep Q-network \cite{Hester2018}. In one curriculum-based approach, the agent learns autonomously, but using demonstration states as starting points \cite{Salimans2018}; this is another effective strategy for Montezuma's Revenge.

Some LfD approaches use demonstration to facilitate decomposition. For example, one method achieves state abstraction by removing features that appear to be irrelevant to demonstrator actions \cite{Cobo2011}. A subsequent method goes on to define subtasks based on differences in their state abstractions \cite{Cobo2012}. Several approaches construct option policies based on segments of demonstrations \cite{Konidaris2011,Fox2017,Zang2009}. In the context of robotics, demonstrations have been used to teach object affordances, which roughly correspond to macro-actions that robots can perform upon objects \cite{Chu2016}.

Decomposing tasks via demonstration is arguably an appealing middle ground between designing subtasks by hand and learning them from scratch. Demonstration often requires less expertise than engineering. And even humans are rarely expected to learn complex tasks without seeing them demonstrated first.

\section{Reasoning from Demonstration}

What would a human learn from a demonstration of Montezuma's Revenge? For an anecdote, I asked a colleague with no history of gaming to watch and describe a YouTube video of the first room. He reported that ``the cowboy had to go get something and bring it up there." Compared to what most LfD agents acquire from demonstration, what my colleague retained seems at once much less and much more. He would likely recall no specific states or actions, but he could explain how to succeed in this task.

This paper asks how an LfD agent could acquire this kind of high-level knowledge from a demonstration, and how it could use that knowledge as it trains. Since no previous LfD systems appear to be based on these questions, this is a fundamental difference between the proposed approach and the existing ones. Let us call this approach Reasoning from Demonstration (RfD).

The goal of an RfD agent is to benefit from demonstration more like humans seem to do. The core of the approach is causal model-building based on objects and their interactions. Given a demonstration, an RfD agent generates a set of cause-effect hypotheses. Causes are object interactions, and effects are other observable events, such as object appearances and environment feedback. Using these hypotheses, an RfD agent can identify desirable and undesirable object interactions. This provides a basis for task decomposition.

The rest of this section describes an RfD agent that illustrates one possible implementation of the broader concept. After motivating the agent's design, the section introduces formal notation and procedures. To facilitate reproduction and future work, all of the code for this agent is available on GitHub\footnote{https://github.com/lisatorrey/reasoning-from-demonstration}, along with all of the data generated in the experiments further below.

\subsection{Objectives and anti-objectives}

Let an \emph{objective} be an object interaction that contributes towards success in the task. For example, in Montezuma's Revenge, the first objective should be to bring the main character, Panama Joe, into contact with the key, causing the door to become openable. The second objective should be to bring Joe into contact with the openable door, causing the success feedback. After observing these events in a demonstration, any RfD agent should form  hypotheses about causes and effects, and these hypotheses should allow it to reason about objectives.

Humans seem to instinctively focus our attention and monitor our progress in object-oriented ways. For example, if we are trying to bring Joe to the key, we will focus primarily on those two objects, and we will be aware that we are making progress as they get closer together. The RfD agent presented in this paper therefore performs object-oriented state abstraction and distance-based reward shaping as it pursues objectives.

Naive distance-based shaping would encourage straight-line movement, which is not always correct. For example, Joe's only viable path to the key nearly circles the entire room. Progress towards objectives needs to be judged along paths that respect the environmental terrain. The RfD agent presented in this paper therefore develops a map of the region connectivity of its environment, so that it can shape rewards effectively.

Let an \emph{anti-objective} be an object interaction that causes failure. For example, the task in Montezuma's Revenge fails if Joe comes into contact with the skull. Since a successful demonstration would not convey this knowledge, it must be discovered during autonomous practice. Any RfD agent should therefore continue to acquire cause-effect hypotheses as it trains.

Anti-objectives need to be avoided during the pursuit of objectives. However, there is neurological evidence that human brains process risks separately from rewards \cite{Xue2009}. The RfD agent presented in this paper therefore develops separate policies with respect to objectives and anti-objectives, and consults both when choosing actions.

\subsection{Objects and events}

Let an \emph{environment} consist of a state space, an action space, and an unknown probability distribution over state transitions. Let a \emph{task} be defined over an environment by choosing four subsets of the state space: the state sets where task attempts begin, end, receive {\small \texttt{SUCCESS}} feedback, and receive {\small \texttt{FAILURE}} feedback. In this task formulation, there are no intermediate rewards.

Any RfD agent needs to be able to perceive objects and events in its environment, but agents may differ in how they represent these elements. Here is the representation used by the RfD agent presented in this paper.

Let {\small \texttt{objects($s$)}} be the set of objects present in state $s$. Given an object, let {\small \texttt{type(object)}} be a descriptor that similar objects would share. Let {\small \texttt{location($s$, object)}} and {\small \texttt{velocity($s$, object)}} describe the perceived position and momentum of the object in state $s$. Let {\small \texttt{region($s$, object)}} indicate which region of the environment the object occupies in state $s$.

Let {\small \texttt{events($s,s'$)}} be the set of object interactions observed at the transition from state $s$ to $s'$. Given an event, let {\small \texttt{type(event)}} be a descriptor that similar events would share. Let {\small \texttt{actor(event)}} be the object  responsible for the event, and let {\small \texttt{subject(event)}}, if it exists, be the object acted upon.

Given an event, let {\small \texttt{template(event)}} be composed of three types: {\small \texttt{type(event)}}, {\small \texttt{type(actor(event))}}, and {\small \texttt{type(subject(event))}}. Let a template be called \emph{instantiable} in states that contain appropriately typed objects to fill the actor and subject roles. Let instantiated templates be called \emph{possible events}. Let {\small \texttt{instances($s$, templates)}} be the set of possible events generated by instantiating one or more templates in state $s$.

Given a possible event, let {\small \texttt{distance($s$, event)}} be the  distance between its objects in state $s$. Let {\small \texttt{focus($s$, event)}} be an abstract state representation for the possible event. In this paper, abstract states consist of the velocities of the objects and the location of the actor (relative to the subject, if there is one) in state $s$.

\subsection{Training procedures}

The main procedure for the RfD agent presented in this paper is Algorithm \ref{RfD}, which augments the standard RL loop for episodic training. The agent develops three kinds of knowledge: a theory, a map, and a set of policies. It begins to develop some of these components based on a demonstrated state sequence, and improves them all as it practices the task autonomously.

A \emph{theory} is a set of cause-effect hypotheses. The agent constructs a theory through repeated application of Algorithm \ref{update-theory}, which reflects a few simple assumptions about causality. It assumes object interactions are potential causes, while object appearances and environment feedback are potential effects. Furthermore, it assumes causes and effects coincide in time; if it observes potential cause $C$ and potential effect $E$ in the same state, it generates the hypothesis $C \longrightarrow E$. Finally, it assumes the rule of logical implication; if it observes $C$ but not $E$, it rejects the hypothesis $C \longrightarrow E$.

\begin{algorithm}[t]
\footnotesize
\begin{algorithmic}
\Procedure{rfd}{}
	\For {each demonstrated transition $(s,s')$}
		\State \texttt{UPDATE-MAP($s$, $s'$)}
		\State \texttt{UPDATE-THEORY($s$, $s'$)}
	\EndFor
	\While {more training is desirable}
		\State $s \longleftarrow$ random choice from initial task states
		\While {$s$ is non-terminal and attempt time < $\tau$}
			\State \texttt{anti-objectives $\longleftarrow$ instances($s$, causes(FAILURE))}
			\State \texttt{objectives $\longleftarrow$ instances($s$, CONTRIBUTORS(SUCCESS))}
			\State \texttt{objective $\longleftarrow$ CHOOSE-OBJECTIVE(objectives)}
			\If {\texttt{objective} is multi-regional}
				\State \texttt{objective $\longleftarrow$ FIRST-CHECKPOINT(objective)}
			\EndIf
			\State $a \longleftarrow$ \texttt{CHOOSE-ACTION($s$, objective, anti-objectives)}
			\State $s' \longleftarrow$ result of taking action $a$
			\State \texttt{UPDATE-MAP($s$, $s'$)}
			\State \texttt{UPDATE-THEORY($s$, $s'$)}
			\State \texttt{UPDATE-POLICIES($s$, $a$, $s'$, objective, anti-objectives)}
			\State $s \longleftarrow s'$
		\EndWhile
	\EndWhile
\EndProcedure
\end{algorithmic}
\caption{Outlines an RfD agent.}
\label{RfD}
\end{algorithm}

\begin{algorithm}[t]
\footnotesize
\begin{algorithmic}
\Procedure{update-theory}{$s$, $s'$}
	\For {each \texttt{event} in \texttt{events($s,s'$)}}
		\If {\texttt{template(event)} has not been seen before}
			\For {each \texttt{object} in \texttt{objects($s'$) $-$ objects($s$)}}
				\State add \texttt{template(event) $\longrightarrow$ type(object)} to the agent's theory
			\EndFor
			\If {\texttt{SUCCESS} occurred in $s'$}
				\State add \texttt{template(event) $\longrightarrow$ SUCCESS} to the agent's theory
			\EndIf
			\If {\texttt{FAILURE} occurred in $s'$}
				\State add \texttt{template(event) $\longrightarrow$ FAILURE} to the agent's theory
			\EndIf
		\EndIf
	\EndFor
	\For {each \texttt{cause $\longrightarrow$ effect} in the agent's theory}
	    \If {\texttt{events($s,s'$)} contains \texttt{event} s.t. \texttt{template(event) = cause}}
		    \If {\texttt{effect} did not occur in $s'$}
				\State remove \texttt{cause $\longrightarrow$ effect} from the agent's theory
			\EndIf
		\EndIf
	\EndFor
\EndProcedure
\end{algorithmic}
\caption{Makes a theory consistent with a transition $(s,s')$.}
\label{update-theory}
\end{algorithm}

\begin{algorithm}[t]
\footnotesize
\begin{algorithmic}
\Procedure{contributors}{$s$, $E$}
	\State \texttt{templates} $\longleftarrow \emptyset$
	\For {each $C$ in \texttt{causes($E$)}}
		\If {$C$ is instantiable in $s$}
			\State add $C$ to \texttt{templates}
		\Else
			\For {each \texttt{object-type} required to make $C$ instantiable in $s$}
				\State merge \texttt{CONTRIBUTORS($s$, object-type)} into templates
			\EndFor
		\EndIf
	\EndFor
	\State \Return templates
\EndProcedure
\end{algorithmic}
\caption{Generates objectives contributing to an effect $E$.}
\label{contributors}
\end{algorithm}

The agent uses its theory to identify objectives and anti-objectives. Anti-objectives are possible events to be avoided, since they may cause failure. Objectives are possible events to be pursued, because they contribute along some causal path towards success, as identified in Algorithm \ref{contributors}. In this algorithm, {\small \texttt{causes($E$)}} indicates the set of all $C$ such that $C \longrightarrow E$ is in the agent's theory.

A \emph{map} is a graph of the regions of the environment and their connectivity. The agent constructs this map through repeated application of Algorithm \ref{update-map}, which keeps track of where objects move across regions. To keep the map small, the agent stores just one transition for each pair of regions.

The agent uses its map to break down multi-region objectives, which involve objects in different regions. It applies Dijkstra's algorithm \cite{Dijkstra1959} to find the shortest path from actor to subject, using region-transitions as intermediate points. The length of a path is the sum of the lengths of its segments. The {\small \texttt{FIRST-CHECKPOINT}} procedure constructs an intermediate objective of moving the actor to the first region-transition along this path. The agent also uses its map whenever there are multiple objectives to choose from; its {\small \texttt{CHOOSE-OBJECTIVE}} procedure selects an objective with minimal path length, breaking ties randomly. (The {\small \texttt{FIRST-CHECKPOINT}} and {\small \texttt{CHOOSE-OBJECTIVE}} subprocedures are the only ones for which pseudocode is omitted due to space constraints.)

A \emph{policy} evaluates actions with respect to a possible event. The agent represents all of its policies with simple Q-functions, and associates one Q-function with each event template. Whenever it discovers a new event template, it creates a new zero-valued Q-function. As it trains, it generates its own rewards to update the appropriate Q-functions, as shown in Algorithm \ref{update-policies}. Objective and checkpoint rewards are based on progress, and all policies have a bonus $\omega$ for completion.

The agent chooses actions by consulting the policies of its current objective and anti-objectives. It uses the corresponding Q-functions to compute both a reward estimate and a cumulative risk estimate for each action, as shown in Algorithm \ref{choose-action}. When it decides to explore, it randomly chooses a minimal-risk action. Otherwise, it chooses an action that maximizes the gap between reward and risk, using a weight $\beta$ to control risk tolerance.

Exploration rates ($\epsilon$) and risk tolerances ($\beta$) are specific to the current objective. New Q-functions for objectives begin with $\epsilon_Q = \epsilon_{max}$. Whenever an objective is completed, its $\epsilon$ decays, with a floor of $\epsilon_{min}$. The agent therefore explores less the more an objective succeeds. New Q-functions for objectives also begin with $\beta_Q = \beta_{max}$. Whenever an objective is exploited, its $\beta$ decays, but whenever it is completed, its $\beta$ resets to $\beta_{max}$. The agent therefore tolerates more risk the longer it pursues an objective without completing it.

All of the experiments in this paper use the following settings for the Greek-letter parameters: $\alpha = 0.1$, $\gamma = 0.9$, $\omega = 100$, $\tau = 10000$, $\epsilon_{max} = 0.1$, $\epsilon_{min} = 0.01$, $\lambda_\epsilon = 0.99$, $\beta_{max} = 100$, $\lambda_\beta = 0.99$.

\begin{algorithm}[t]
\footnotesize
\begin{algorithmic}
\Procedure{update-map}{$s$, $s'$}
	\For {each \texttt{object} in \texttt{objects($s$) $\cap$ objects($s'$)}}
		\State $R \longleftarrow$ \texttt{region($s$, object)}
		\State $R' \longleftarrow$ \texttt{region($s'$, object)}
		\If {$R \neq R'$ and no transition $R \longrightarrow R'$ exists in the agent's map}
			\State add transition $R \longrightarrow R'$ to the agent's map at \texttt{location($s'$, object)}
		\EndIf
	\EndFor
\EndProcedure
\end{algorithmic}
\caption{Adds to a map based on a state transition $(s,s')$.}
\label{update-map}
\end{algorithm}

\begin{algorithm}[t]
\footnotesize
\begin{algorithmic}
\Procedure{update-policies}{$s$, $a$, $s'$, \texttt{objective}, \texttt{anti-objectives}}
	\State $Q \longleftarrow$ the Q-function associated with \texttt{template(objective)}
	\State $\delta \longleftarrow$ \texttt{distance($s$, objective) $-$ distance($s'$, objective)}
	\State $s \longleftarrow$ \texttt{focus($s$, objective)}
	\State $s' \longleftarrow$ \texttt{focus($s'$, objective)}
	\If {\texttt{objective} is a checkpoint}
		\If {\texttt{actor(objective)} has entered the checkpoint region}
			\State update $Q$ according to Equation 1 with $s$, $a$, $s'$, and $r = \delta + \omega$
			\State $\epsilon_Q \longleftarrow \max(\epsilon_{min}, \lambda_\epsilon \epsilon_Q)$
			\State $\beta_Q \longleftarrow \beta_{max}$
		\ElsIf {\texttt{actor(objective)} has entered a different region}
			\State update $Q$ according to Equation 1 with $s$, $a$, $s'$, and $r = \delta - \omega$
		\Else
			\State update $Q$ according to Equation 1 with $s$, $a$, $s'$, and $r = \delta$
		\EndIf
	\Else
		\If {\texttt{objective $\in$ events($s,s'$)}}
			\State update $Q$ according to Equation 1 with $s$, $a$, $s'$, and $r = \delta + \omega$
			\State $\epsilon_Q \longleftarrow \max(\epsilon_{min}, \lambda_\epsilon \epsilon_Q)$
			\State $\beta_Q \longleftarrow \beta_{max}$
		\ElsIf {\texttt{objective} remains possible in $s'$}
			\State update $Q$ according to Equation 1 with $s$, $a$, $s'$, and $r = \delta$
		\EndIf
	\EndIf
	\For {each \texttt{anti-objective}}
		\State $Q \longleftarrow$ the Q-function associated with \texttt{template(anti-objective)}
		\State $s$ $\longleftarrow$ \texttt{focus($s$, anti-objective)}
		\State $s' \longleftarrow$ \texttt{focus($s'$, anti-objective)}
		\If {\texttt{anti-objective $\in$ events($s,s'$)}}
			\State update $Q$ according to Equation 1 with $s$, $a$, $s'$, and $r = -\omega$
		\ElsIf {\texttt{anti-objective} remains possible in $s'$}
			\State update $Q$ according to Equation 1 with $s$, $a$, $s'$, and $r = 0$
		\EndIf
	\EndFor
\EndProcedure
\end{algorithmic}
\caption{Adjusts policies after a transition $(s,a,s')$.}
\label{update-policies}
\end{algorithm}

\begin{algorithm}[t]
\footnotesize
\begin{algorithmic}
\Procedure{choose-action}{$s$, \texttt{objective}, \texttt{anti-objectives}}
	\For {each $a$ in the action space}
		\State \texttt{risk[$a$] $\longleftarrow 0$}
	\EndFor
	\For {each \texttt{anti-objective}}
		\State $Q \longleftarrow$ the Q-function associated with \texttt{template(anti-objective)}
		\State $s \longleftarrow$ \texttt{focus($s$, anti-objective)}
		\State \texttt{risk[$a$] $\longleftarrow$ risk[$a$] $-$ $Q(s,a)$}
	\EndFor
	\State $Q$ $\longleftarrow$ the Q-function associated with \texttt{template(objective)}
	\State $s$ $\longleftarrow$ \texttt{focus($s$, objective)}
	\For {each $a$ in the action space}
		\State \texttt{reward[$a$] $\longleftarrow$ $Q(s,a)$}
	\EndFor
	\If {\texttt{random(0,1)} $ < \epsilon_Q$}
		\State \texttt{safest} $\longleftarrow$ set of $a$ minimizing \texttt{risk[$a$]}
		\State \Return random choice from \texttt{safest}
	\Else
		\State \texttt{best} $\longleftarrow$ set of $a$ maximizing \texttt{reward[$a$] $-$ $\beta_Q$ risk[$a$]}
		\State $\beta_Q \longleftarrow \lambda_\beta \beta_Q$
		\State \Return random choice from \texttt{best}
	\EndIf
\EndProcedure
\end{algorithmic}
\caption{Chooses an action in state $s$.}
\label{choose-action}
\end{algorithm}

\section{Benchmark}

Taxi makes a good benchmark for comparing RfD with LfD. It is small enough to apply any approach, but sparse enough to reveal differences between approaches. There are two main categories of LfD that should be considered: approaches that use demonstration to influence a single policy, and approaches that use demonstration to decompose the task. This section identifies one approach in each category that seems best suited to the Taxi problem, and shows how these two LfD approaches compare with RfD.

It is easy to generate good demonstrations for Taxi. Training a standard Q-learner to convergence takes about 100,000 actions, and then it makes a reliable demonstrator. However, a Taxi demonstration provides very limited information. There are 500 unique states in the environment, and good demonstrations encounter only 5 to 20 of them. Furthermore, the traditional state representation provides little opportunity to generalize beyond a demonstration. States that match in some or even most of their feature values still often have different optimal actions.

\begin{figure}
\centering
\includegraphics[width=\columnwidth]{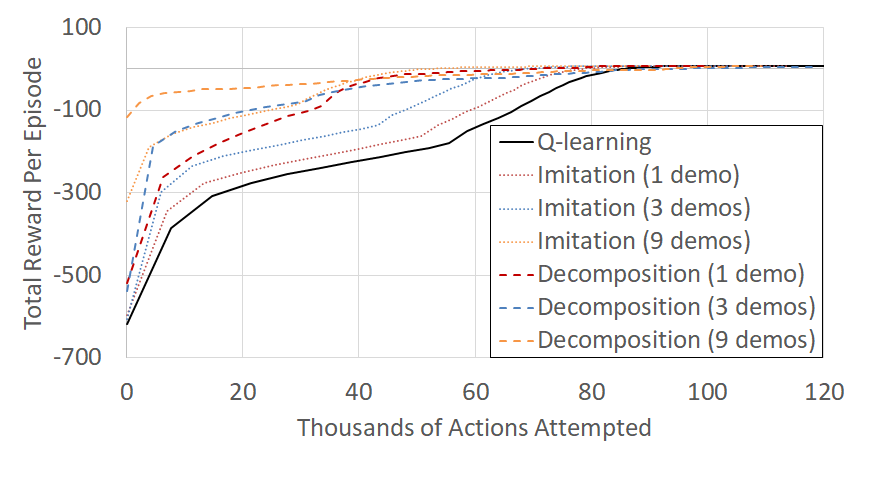}
\caption{\small{Learning curves for non-RfD agents in Taxi. Curves are smoothed over 400 episodes and averaged over 10 different agents.}}
\label{imitation}
\end{figure}

In this situation, a single-policy LfD agent should benefit most by favoring imitation over generalization. The imitation agents in this experiment therefore use standard Q-learning, but always take demonstrated actions in demonstrated states, unless they are exploring. Figure \ref{imitation} confirms that demonstrations are useful in Taxi: the imitation agents learn faster than the standard Q-learner, and the effect increases with the number of demonstrations.

Taxi does provide opportunities for task decomposition, as well as further state decomposition within subtasks. Among the existing approaches, Automatic Decomposition and Abstraction (ADA) seems best suited to take advantage of both opportunities \cite{Cobo2012}.

ADA partitions a state space into subtasks along numeric feature boundaries, and removes irrelevant features in subtasks based on their mutual information with demonstrator actions. In Taxi, the {\small \texttt{passenger}} feature has values in $\{0,1,2,3,4\}$, where 4 means the passenger is inside the taxi, and the other values represent the stops. Thus it is appropriate to use the boundary {\small \texttt{passenger}} $< 4$ to divide Taxi into two subtasks. One side is the pickup subtask, in which the {\small \texttt{destination}} feature should be ignored. The other side is the dropoff subtask, in which the {\small \texttt{passenger}} feature can be ignored.

In 100 trials of up to 32 demonstrations generated one at a time by a trained Q-learner, ADA produced the correct decomposition 72 times, after a minimum of 3 demonstrations and an average of 12. To establish an upper bound on the potential of ADA, the decomposition agents in this experiment all use the correct decomposition, and they train the two subtasks via imitation. Figure \ref{imitation} confirms that decomposition is useful in Taxi: given the same number of demonstrations, decomposition agents quickly outperform imitation agents.

The RfD agents each receive just one demonstration, which I performed from the initial state shown in Figure \ref{tasks}. Instead of the usual numeric rewards of the Taxi environment, the RfD agents receive only {\small \texttt{SUCCESS}} or {\small \texttt{FAILURE}} feedback at the end of each attempt. The regions of the environment are the five rectangles suggested by the walls. Each object has a {\small \texttt{(row, column)}} location and a velocity of 0. The object types and event templates are evident from these hypotheses, which the RfD agents generate for Taxi:

\vspace{1mm}
\footnotesize
\begin{center}
\begin{tabular}{l}
\hline
picks(Taxi, Passenger) $\longrightarrow$ Taxi+Passenger \\
picks(Taxi, Passenger) $\longrightarrow$ Stop \\
\hline
drops(Taxi+Passenger, Stop) $\longrightarrow$ Taxi \\
drops(Taxi+Passenger, Stop) $\longrightarrow$ Passenger \\
\hline
drops(Taxi+Passenger, Destination) $\longrightarrow$ \texttt{SUCCESS} \\
drops(Taxi+Passenger, Destination) $\longrightarrow$ Stop \\
\hline
\end{tabular}
\end{center}
\normalsize
\vspace{1mm}

\begin{figure}
\centering
\includegraphics[width=\columnwidth]{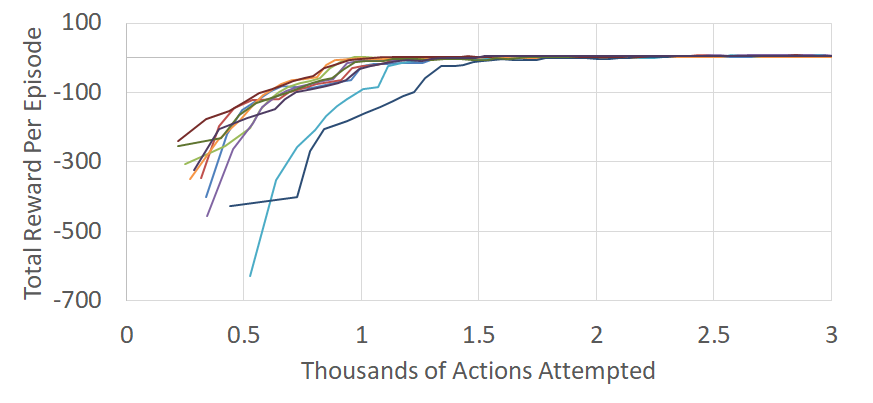}
\caption{\small{Learning curves for ten RfD agents in Taxi. Curves are 300 attempts long and smoothed over a 30-attempt window.}}
\label{taxi-curves}
\end{figure}

Differences in the forms and amounts of assistance that the RfD and decomposition agents receive complicate the comparison of their learning curves. That said, the differences in their learning curves are dramatic. The RfD agents converge in about 2\% of the time that the decomposition agents take, regardless of the number of demonstrations. Figure \ref{taxi-curves} shows the RfD learning curves on an appropriate scale.

Although they represent the information differently, the RfD and decomposition agents are both solving the same subtasks in the same state space. The difference in their performance is mainly attributable to the reward shaping that the RfD agents perform. Of course, it is no surprise that reward shaping speeds up learning. But agents need causal and spatial knowledge in order to perform reward shaping themselves, without human engineering.

Overall, the comparison of ADA and RfD suggests that decomposition based on causal reasoning, rather than statistical reasoning, has the potential to facilitate significantly faster learning after significantly less demonstration.

\section{Challenges}

This section shows how RfD scales to larger and sparser environments: Courier, Ms. Pacman, and Montezuma's Revenge.

\subsection{Courier}

With four packages and 20 vehicles on a 35x35 grid, just the \emph{initial} states of the Courier task number more than $10^{35}$. My single demonstration begins at the state shown in Figure \ref{tasks}, collects the two packages on one side, collects the two packages on the other side, and delivers them all at once.

The regions of the environment are the three rectangles suggested by the walls. Each object has a {\small \texttt{(row, column)}} location. The object types and event templates are evident from these initial hypotheses that the RfD agents generate for Courier:

\vspace{1mm}
\footnotesize
\begin{center}
\begin{tabular}{l}
\hline
arrives(Courier, Package) $\longrightarrow$ Courier+1 \\
arrives(Courier+1, Package) $\longrightarrow$ Courier+2 \\
arrives(Courier+2, Package) $\longrightarrow$ Courier+3 \\
arrives(Courier+3, Package) $\longrightarrow$ Courier+4 \\
\hline
arrives(Courier+4, Platform) $\longrightarrow$ \texttt{SUCCESS} \\
arrives(Courier+4, Platform) $\longrightarrow$ Platform+4 \\
arrives(Courier+4, Platform) $\longrightarrow$ Courier \\
\hline
\end{tabular}
\end{center}
\normalsize
\vspace{1mm}

Agents add additional hypotheses as they train. Most importantly, they discover that collides(Courier$+k$, Vehicle) causes {\small \texttt{FAILURE}} for any $k$. They also tend to discover the possibility of delivering fewer than four packages at once. These discoveries lead some agents to develop alternate strategies.

\begin{figure}
\centering
\includegraphics[width=\columnwidth]{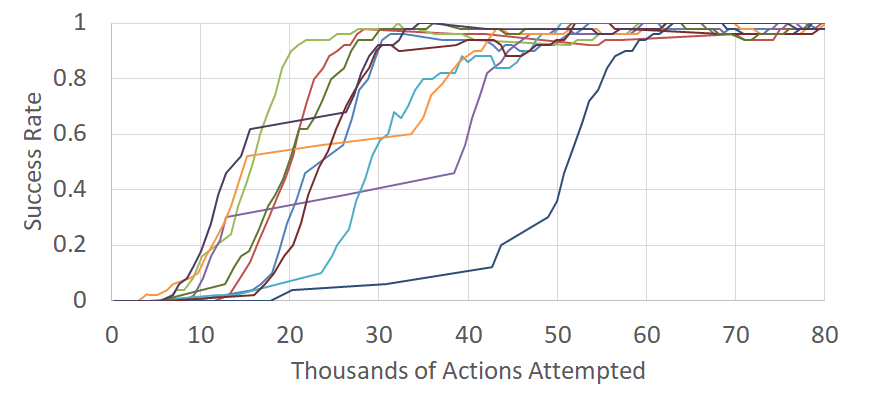}
\caption{\small{Learning curves for ten RfD agents in Courier. Curves are 500 attempts long and smoothed over a 50-attempt window.}}
\label{courier-curves}
\end{figure}

The ADA approach is not practically applicable in Courier due to the dimensionality of the state space. And because Courier is essentially designed to need object-oriented decomposition, no other LfD approach seems more applicable. However, Figure \ref{courier-curves} shows that training RfD agents to convergence in Courier takes less than 100,000 actions. These results indicate that combinatorially huge environment sizes and arbitrarily long subtask sequences are surmountable via RfD. Furthermore, they show that RfD agents can develop logical solutions that were not demonstrated.

\subsection{Ms. Pacman}

In Ms. Pacman, agents can accumulate rewards through a sequence of levels just by eating densely placed pellets. Regular pellets are worth 10 points, while the rarer power-pellets are worth 50 points. However, the highest rewards are only available for a short period after consuming a power-pellet, when the ghosts become temporarily edible. Catching one turns it into a pair of eyes and yields 200 points, and this reward doubles for each additional catch during the same period.

Let us define a task in Ms. Pacman that focuses on catching edible ghosts in the first level. It provides {\small \texttt{SUCCESS}} feedback when Ms. Pacman catches an edible ghost and {\small \texttt{FAILURE}} feedback when she collides with a regular ghost. The task ends upon failure, but success is achievable up to 16 times per attempt, because there are four power-pellets and four ghosts.

This game always begins in the same state, and the Atari emulator is, unfortunately, inherently deterministic; the same sequence of player actions always produces the same sequence of states. A common way to introduce some stochasticity is to delay the start of the game, allowing the ghosts to move for a random number of frames before Ms. Pacman can start moving.

The environment provides only pixels, but objects in Ms. Pacman can be identified based on their colors. Each object has an {\small \texttt{(x,y)}} location that approximates its center. Since the typical practice in Atari games is to repeat actions for four frames each, any moving object has a velocity, which is expressed as {\small \texttt{UP}}, {\small \texttt{DOWN}}, {\small \texttt{LEFT}}, or {\small \texttt{RIGHT}}. The object types and event templates are evident from these hypotheses, which the RfD agents generate for Ms. Pacman:

\vspace{1mm}
\footnotesize
\begin{center}
\begin{tabular}{l}
\hline
arrives(Pacman, Power) $\longrightarrow$ Edible \\
\hline
catches(Pacman, Edible) $\longrightarrow$ \texttt{SUCCESS} \\
catches(Pacman, Edible) $\longrightarrow$ Eyes \\
\hline
collides(Pacman, Ghost) $\longrightarrow$ \texttt{FAILURE} \\
\hline
\end{tabular}
\end{center}
\normalsize
\vspace{1mm}

Demonstrations for this task could vary widely. I performed a short one, taking Ms. Pacman directly to the lower left pellet and keeping her in that corner, where one edible ghost soon stumbles into her, and a bit later a regular ghost finds her. Unlike my demonstrations for other tasks, this one provides the basis for a complete causal model, but for only a partial map of the regions (which are the 36 corridors). The RfD agents discover about 60\% of the regions and about 80\% of the transitions as they train.

\begin{figure}
\centering
\includegraphics[width=\columnwidth]{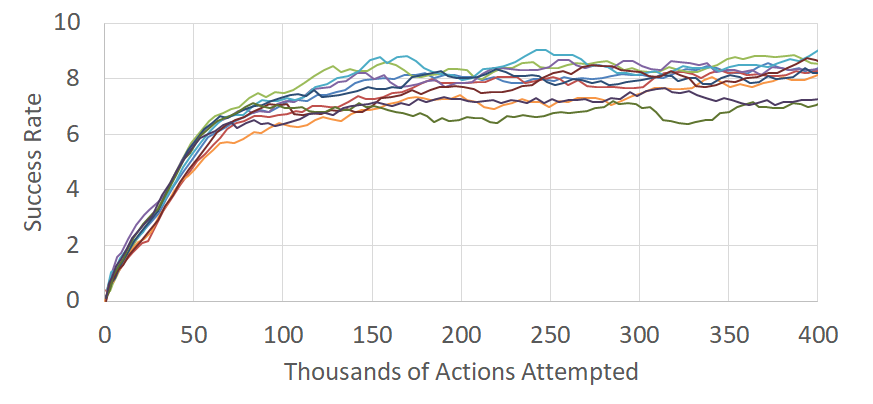}
\caption{\small{Learning curves for ten RfD agents in Ms. Pacman. Curves are 1000 attempts long and smoothed over a 100-attempt window.}}
\label{pacman-curves}
\end{figure}

Figure \ref{pacman-curves} shows that navigation-intensive problems are approachable via RfD. The RfD agents learn to average about 8 successes per attempt within a few hundred thousand actions.
Since most agents in Ms. Pacman focus on maximizing points across multiple levels, comparisons to previous results are not necessarily meaningful. However, one agent that was heavily engineered to maximize its score on the first level of Ms. Pacman averaged about 3800 points at its asymptote \cite{Torrey2013}, while at their asymptotes, the RfD agents average about 5600 points on that level. Causal modeling allows them to pursue large rewards in more deliberate ways. 

\subsection{Montezuma's Revenge}

The first-room task in Montezuma's Revenge provides {\small \texttt{SUCCESS}} feedback when Joe exits the room and {\small \texttt{FAILURE}} feedback when he loses a life. In both cases, the task ends. 

Regions correspond to the visual elements of the terrain: the platforms, ladders, floor, and rope. Objects are identified based on their colors, and each object has an $(x,y$) location that approximates its center. Moving objects have velocities that are expressed as $(\Delta_x, \Delta_y)$ tuples. The object types and event templates are evident from these  hypotheses, which the RfD agents generate for Montezuma's Revenge:

\vspace{1mm}
\footnotesize
\begin{center}
\begin{tabular}{l}
\hline
arrives(Joe, Key) $\longrightarrow$ Door+Key \\
arrives(Joe, Door+Key) $\longrightarrow$ \texttt{SUCCESS} \\
\hline
collides(Joe, Skull) $\longrightarrow$ \texttt{FAILURE} \\
falls(Joe) $\longrightarrow$ \texttt{FAILURE} \\
\hline
\end{tabular}
\end{center}
\normalsize
\vspace{1mm}

There is little variation among successful demonstrations for this task, because they must all take approximately the same route down to the key and back up to the door. To introduce some variation, the start of each game is randomly delayed by up to 400 frames, so that Joe might reach the skull at any point in its cyclical patrol. My demonstration provides the basis for a complete map, but only a partial theory; the RfD agents must learn the causes of failure as they train.

\begin{figure}
\centering
\includegraphics[width=\columnwidth]{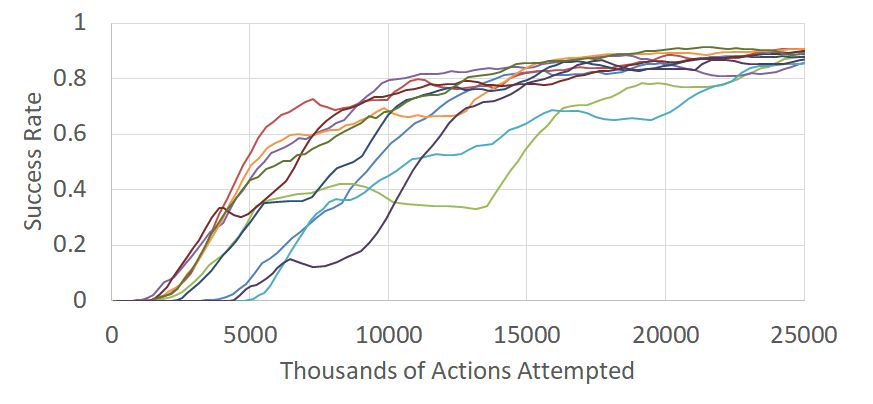}
\caption{\small{Learning curves for ten RfD agents in Montezuma's Revenge. Curves are 100,000 attempts long and smoothed over a 10,000-attempt window.}}
\label{montezuma-curves}
\end{figure}

Figure \ref{montezuma-curves} shows that the RfD agents reach a success rate of about 90\% in this task after about 25 million actions. Comparisons to previous results are not necessarily meaningful, because most agents in Montezuma's Revenge are permitted to use all five lives. Since the skull disappears after one collision, even having two lives to spend makes it much easier to exit the room.

One approach called Deep Abstract Q-Networks does consider the single-life condition \cite{Roderick2018}. Using hierarchical deep RL with human-engineered abstract states, it rises to a peak average of approximately 300 points in approximately 50 million frames. These results seem comparable to the RfD results. The similarity suggests that when learning sparse-reward tasks, it may be less important to use a sophisticated RL algorithm than it is to acquire, in some way, an effective decomposition.

\subsection{Extended actions}

The task in Montezuma's Revenge is only slightly longer than the other tasks examined in this paper. The difference is far too small to explain the steep increase in learning time. Why is Montezuma's Revenge so hard to learn, even for agents who are well-equipped to handle sparse rewards? One recent study notes that the environment contains many dead ends, or states from which an agent will certainly lose a life, despite continuing to act for a while before it occurs \cite{Fatemi2019}. For example, if Joe steps off the top platform, he falls to his death, and although the emulator continues to accept actions on the way down, none of them have any effect.

From an RL agent's perspective, this problem can be characterized in a different way: actions in Montezuma's Revenge have unpredictable durations. Normally each action lasts for four frames, but if an action begins a jump or a fall, it really lasts until Joe returns to a surface again. Humans understand this fact intuitively, since our experience in the physical world makes us expect to be powerless to change our trajectory in mid-air. Because of our intuition, we perceive the actions in this environment differently than RL agents do.

\begin{figure}[t]
\centering
\includegraphics[width=\columnwidth]{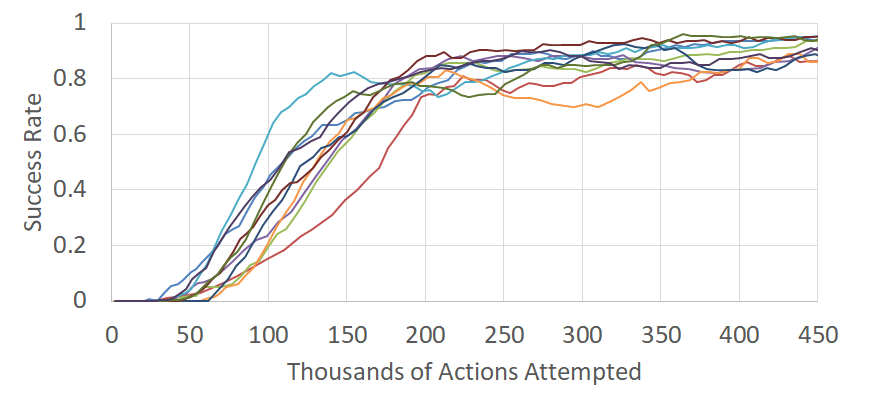}
\caption{\small{Learning curves for ten RfD agents in Montezuma's Revenge with extended actions. Curves are 2000 attempts long and smoothed over a 200-attempt window.}}
\label{extended-curves}
\end{figure}

Figure \ref{extended-curves} shows that this difference in perception has a dramatic effect on the learning process. If the actions in Montezuma's Revenge are extended until Joe stops falling, to correspond with human perceptions, the RfD agents can reach the same asymptote in about 2\% of the time. The knowledge that RfD agents develop is insufficient for them to make this adjustment autonomously, so these results should not be considered achievements of RfD. They are included here just to clarify the difficulty of Montezuma's Revenge, which is due not only to the nature of its rewards, but also to the nature of its actions.

\section{Conclusions}

This paper proposes a framework called Reasoning from Demonstration. It is inspired by the way that humans seem to approach sparse-reward tasks with object-oriented causal reasoning. Agents in this framework acquire causal knowledge from demonstration and use it to decompose tasks effectively and practice them deliberately. The paper presents experimental evidence that in sparse-reward tasks, RfD approaches have the potential for much faster learning than traditional LfD approaches.

Another potential benefit of RfD agents is their natural resilience to low-quantity and low-quality demonstrations. One successful demonstration is often enough to establish the causal dynamics of a task. And as long as a demonstration ultimately reveals those causal dynamics, it can be arbitrarily sub-optimal without having any detrimental effect on RfD agents.

The agent described in this paper implements the RfD concept using simple components and procedures. Many of them could clearly be improved by other agents within the RfD framework. For example, richer representations of objects and policies would allow for finer control, and a more sophisticated treatment of causality would produce more robust reasoning.

Most importantly, although the agent in this paper is assisted in perceiving objects, events, and regions, future RfD agents could approach these problems autonomously. In fact, the main conclusion of this paper is that research on  perception should be better integrated into research on learning in sparse-reward tasks. Structured perception is necessary for causal modeling, which clearly has potential benefits for sparse-reward learning.

This work joins ongoing discussions about combining statistical and symbolic AI \cite{Dietterich2019} and about making agents learn more like humans do \cite{Lake2017}. These discussions seem particularly relevant to the area of LfD, which is already human-inspired.

\bibliographystyle{ACM-Reference-Format}
\bibliography{rfd}


\begin{thebibliography}{00}


\ifx \showCODEN    \undefined \def \showCODEN     #1{\unskip}     \fi
\ifx \showDOI      \undefined \def \showDOI       #1{#1}\fi
\ifx \showISBNx    \undefined \def \showISBNx     #1{\unskip}     \fi
\ifx \showISBNxiii \undefined \def \showISBNxiii  #1{\unskip}     \fi
\ifx \showISSN     \undefined \def \showISSN      #1{\unskip}     \fi
\ifx \showLCCN     \undefined \def \showLCCN      #1{\unskip}     \fi
\ifx \shownote     \undefined \def \shownote      #1{#1}          \fi
\ifx \showarticletitle \undefined \def \showarticletitle #1{#1}   \fi
\ifx \showURL      \undefined \def \showURL       {\relax}        \fi
\providecommand\bibfield[2]{#2}
\providecommand\bibinfo[2]{#2}
\providecommand\natexlab[1]{#1}
\providecommand\showeprint[2][]{arXiv:#2}

\bibitem[\protect\citeauthoryear{Bacon, Harb, and Precup}{Bacon
  et~al\mbox{.}}{2017}]%
        {Bacon2017}
\bibfield{author}{\bibinfo{person}{Pierre-Luc Bacon}, \bibinfo{person}{Jean
  Harb}, {and} \bibinfo{person}{Doina Precup}.}
  \bibinfo{year}{2017}\natexlab{}.
\newblock \showarticletitle{{The Option-Critic Architecture}}. In
  \bibinfo{booktitle}{{\em Proceedings of the 31st AAAI Conference on
  Artificial Intelligence}}.
\newblock


\bibitem[\protect\citeauthoryear{Brys, Harutyunyan, Suay, Chernova, Taylor, and
  Now{\'e}}{Brys et~al\mbox{.}}{2015}]%
        {Brys2015}
\bibfield{author}{\bibinfo{person}{Tim Brys}, \bibinfo{person}{Anna
  Harutyunyan}, \bibinfo{person}{Halit~Bener Suay}, \bibinfo{person}{Sonia
  Chernova}, \bibinfo{person}{Matthew Taylor}, {and} \bibinfo{person}{Ann
  Now{\'e}}.} \bibinfo{year}{2015}\natexlab{}.
\newblock \showarticletitle{{Reinforcement Learning from Demonstration Through
  Shaping}}. In \bibinfo{booktitle}{{\em Proceedings of the 24th Joint
  International Conference on Artificial Intelligence}}.
\newblock


\bibitem[\protect\citeauthoryear{Chu, Akgun, and Thomaz}{Chu
  et~al\mbox{.}}{2016}]%
        {Chu2016}
\bibfield{author}{\bibinfo{person}{Vivian Chu}, \bibinfo{person}{Baris Akgun},
  {and} \bibinfo{person}{Andrea Thomaz}.} \bibinfo{year}{2016}\natexlab{}.
\newblock \showarticletitle{{Learning Haptic Affordances from Demonstration and
  Human-Guided Exploration}}. In \bibinfo{booktitle}{{\em Proceedings of the
  IEEE Haptics Symposium}}.
\newblock


\bibitem[\protect\citeauthoryear{Cobo, Isbell, and Thomaz}{Cobo
  et~al\mbox{.}}{2012}]%
        {Cobo2012}
\bibfield{author}{\bibinfo{person}{Luis Cobo}, \bibinfo{person}{Charles
  Isbell}, {and} \bibinfo{person}{Andrea Thomaz}.}
  \bibinfo{year}{2012}\natexlab{}.
\newblock \showarticletitle{{Automatic Task Decomposition and State Abstraction
  from Demonstration}}. In \bibinfo{booktitle}{{\em Proceedings of the 11th
  International Conference on Autonomous Agents and Multiagent Systems}}.
\newblock


\bibitem[\protect\citeauthoryear{Cobo, Isbell, and Thomaz}{Cobo
  et~al\mbox{.}}{2013}]%
        {Cobo2013}
\bibfield{author}{\bibinfo{person}{Luis Cobo}, \bibinfo{person}{Charles
  Isbell}, {and} \bibinfo{person}{Andrea Thomaz}.}
  \bibinfo{year}{2013}\natexlab{}.
\newblock \showarticletitle{{Object Focused Q-learning for Autonomous Agents}}.
  In \bibinfo{booktitle}{{\em Proceedings of the 12th International Conference
  on Autonomous Agents and Multi-agent Systems}}.
\newblock


\bibitem[\protect\citeauthoryear{Cobo, Zang, Isbell, and Thomaz}{Cobo
  et~al\mbox{.}}{2011}]%
        {Cobo2011}
\bibfield{author}{\bibinfo{person}{Luis Cobo}, \bibinfo{person}{Peng Zang},
  \bibinfo{person}{Charles Isbell}, {and} \bibinfo{person}{Andrea Thomaz}.}
  \bibinfo{year}{2011}\natexlab{}.
\newblock \showarticletitle{{Automatic State Abstraction from Demonstration}}.
  In \bibinfo{booktitle}{{\em Proceedings of the 22nd International Joint
  Conference on Artificial Intelligence}}.
\newblock


\bibitem[\protect\citeauthoryear{Dann, Zambetta, and Thangarajah}{Dann
  et~al\mbox{.}}{2019}]%
        {Dann2019}
\bibfield{author}{\bibinfo{person}{Michael Dann}, \bibinfo{person}{Fabio
  Zambetta}, {and} \bibinfo{person}{John Thangarajah}.}
  \bibinfo{year}{2019}\natexlab{}.
\newblock \showarticletitle{{Deriving Subgoals Autonomously to Accelerate
  Learning in Sparse Reward Domains}}. In \bibinfo{booktitle}{{\em Proceedings
  of the 33rd AAAI Conference on Artificial Intelligence}}.
\newblock


\bibitem[\protect\citeauthoryear{Dayan and Hinton}{Dayan and Hinton}{1992}]%
        {Dayan1992}
\bibfield{author}{\bibinfo{person}{Peter Dayan} {and} \bibinfo{person}{Geoffrey
  Hinton}.} \bibinfo{year}{1992}\natexlab{}.
\newblock \showarticletitle{{Feudal Reinforcement Learning}}. In
  \bibinfo{booktitle}{{\em Advances in Neural Information Processing Systems
  5}}.
\newblock


\bibitem[\protect\citeauthoryear{Dietterich}{Dietterich}{2000}]%
        {Dietterich2000}
\bibfield{author}{\bibinfo{person}{Thomas Dietterich}.}
  \bibinfo{year}{2000}\natexlab{}.
\newblock \showarticletitle{{Hierarchical Reinforcement Learning with the MAXQ
  Value Function Decomposition}}.
\newblock \bibinfo{journal}{{\em Journal of Artificial Intelligence
  Research\/}} \bibinfo{volume}{13}, \bibinfo{number}{1}
  (\bibinfo{year}{2000}), \bibinfo{pages}{227--303}.
\newblock


\bibitem[\protect\citeauthoryear{Dietterich}{Dietterich}{2019}]%
        {Dietterich2019}
\bibfield{author}{\bibinfo{person}{Thomas Dietterich}.}
  \bibinfo{year}{2019}\natexlab{}.
\newblock   (\bibinfo{date}{10} \bibinfo{year}{2019}).
\newblock
\showURL{%
\url{https://medium.com/@tdietterich/what-does-it-mean-for-a-machine-to-understand-555485f3ad40}}


\bibitem[\protect\citeauthoryear{Dijkstra}{Dijkstra}{1959}]%
        {Dijkstra1959}
\bibfield{author}{\bibinfo{person}{Edsger Dijkstra}.}
  \bibinfo{year}{1959}\natexlab{}.
\newblock \showarticletitle{{A note on two problems in connexion with graphs}}.
\newblock \bibinfo{journal}{{\it Numer. Math.}}  \bibinfo{volume}{1}
  (\bibinfo{year}{1959}), \bibinfo{pages}{269–--271}.
\newblock


\bibitem[\protect\citeauthoryear{Diuk, Cohen, and Littman}{Diuk
  et~al\mbox{.}}{2008}]%
        {Diuk2008}
\bibfield{author}{\bibinfo{person}{Carlos Diuk}, \bibinfo{person}{Andre Cohen},
  {and} \bibinfo{person}{Michael Littman}.} \bibinfo{year}{2008}\natexlab{}.
\newblock \showarticletitle{{An Object-oriented Representation for Efficient
  Reinforcement Learning}}. In \bibinfo{booktitle}{{\em Proceedings of the 25th
  International Conference on Machine Learning}}.
\newblock


\bibitem[\protect\citeauthoryear{Dubey, Agrawal, Pathak, Griffiths, and
  Efros}{Dubey et~al\mbox{.}}{2018}]%
        {Dubey2018}
\bibfield{author}{\bibinfo{person}{Rachit Dubey}, \bibinfo{person}{Pulkit
  Agrawal}, \bibinfo{person}{Deepak Pathak}, \bibinfo{person}{Thomas
  Griffiths}, {and} \bibinfo{person}{Alexei Efros}.}
  \bibinfo{year}{2018}\natexlab{}.
\newblock \showarticletitle{{Investigating Human Priors for Playing Video
  Games}}. In \bibinfo{booktitle}{{\em Proceedings of the 35th International
  Conference on Machine Learning}}.
\newblock


\bibitem[\protect\citeauthoryear{D{\v{z}}eroski, De~Raedt, and
  Driessens}{D{\v{z}}eroski et~al\mbox{.}}{2001}]%
        {Dzeroski2001}
\bibfield{author}{\bibinfo{person}{Sa{\v{s}}o D{\v{z}}eroski},
  \bibinfo{person}{Luc De~Raedt}, {and} \bibinfo{person}{Kurt Driessens}.}
  \bibinfo{year}{2001}\natexlab{}.
\newblock \showarticletitle{{Relational Reinforcement Learning}}.
\newblock \bibinfo{journal}{{\em Machine Learning\/}} \bibinfo{volume}{43},
  \bibinfo{number}{1} (\bibinfo{year}{2001}), \bibinfo{pages}{7--52}.
\newblock


\bibitem[\protect\citeauthoryear{Fatemi, Sharma, van Seijen, and Kahou}{Fatemi
  et~al\mbox{.}}{2019}]%
        {Fatemi2019}
\bibfield{author}{\bibinfo{person}{Mehdi Fatemi}, \bibinfo{person}{Shikhar
  Sharma}, \bibinfo{person}{Harm van Seijen}, {and}
  \bibinfo{person}{Samira~Ebrahimi Kahou}.} \bibinfo{year}{2019}\natexlab{}.
\newblock \showarticletitle{{Dead-ends and Secure Exploration in Reinforcement
  Learning}}. In \bibinfo{booktitle}{{\em Proceedings of the 36th International
  Conference on Machine Learning}}.
\newblock


\bibitem[\protect\citeauthoryear{Fox, Krishnan, Stoica, and Goldberg}{Fox
  et~al\mbox{.}}{2017}]%
        {Fox2017}
\bibfield{author}{\bibinfo{person}{Roy Fox}, \bibinfo{person}{Sanjay Krishnan},
  \bibinfo{person}{Ion Stoica}, {and} \bibinfo{person}{Ken Goldberg}.}
  \bibinfo{year}{2017}\natexlab{}.
\newblock \showarticletitle{{Multi-Level Discovery of Deep Options}}.
\newblock \bibinfo{journal}{{\em CoRR\/}}  \bibinfo{volume}{abs/1703.08294}
  (\bibinfo{year}{2017}).
\newblock


\bibitem[\protect\citeauthoryear{Garnelo, Arulkumaran, and Shanahan}{Garnelo
  et~al\mbox{.}}{2016}]%
        {Garnelo2016}
\bibfield{author}{\bibinfo{person}{Marta Garnelo}, \bibinfo{person}{Kai
  Arulkumaran}, {and} \bibinfo{person}{Murray Shanahan}.}
  \bibinfo{year}{2016}\natexlab{}.
\newblock \showarticletitle{{Towards Deep Symbolic Reinforcement Learning}}.
\newblock \bibinfo{journal}{{\em arXiv\/}}  \bibinfo{volume}{1609.05518}
  (\bibinfo{year}{2016}).
\newblock


\bibitem[\protect\citeauthoryear{Grollman and Billard}{Grollman and
  Billard}{2012}]%
        {Grollman2012}
\bibfield{author}{\bibinfo{person}{D. Grollman} {and} \bibinfo{person}{A.
  Billard}.} \bibinfo{year}{2012}\natexlab{}.
\newblock \showarticletitle{{Robot Learning from Failed Demonstrations}}.
\newblock \bibinfo{journal}{{\em International Journal of Social Robotics\/}}
  \bibinfo{volume}{4} (\bibinfo{year}{2012}), \bibinfo{pages}{331–342}.
\newblock


\bibitem[\protect\citeauthoryear{Hester, Vecerik, Pietquin, Lanctot, Schaul,
  Piot, Horgan, Quan, Sendonaris, Osband, Dulac{-}Arnold, Agapiou, Leibo, and
  Gruslys}{Hester et~al\mbox{.}}{2018}]%
        {Hester2018}
\bibfield{author}{\bibinfo{person}{Todd Hester}, \bibinfo{person}{Matej
  Vecerik}, \bibinfo{person}{Olivier Pietquin}, \bibinfo{person}{Marc Lanctot},
  \bibinfo{person}{Tom Schaul}, \bibinfo{person}{Bilal Piot},
  \bibinfo{person}{Dan Horgan}, \bibinfo{person}{John Quan},
  \bibinfo{person}{Andrew Sendonaris}, \bibinfo{person}{Ian Osband},
  \bibinfo{person}{Gabriel Dulac{-}Arnold}, \bibinfo{person}{John Agapiou},
  \bibinfo{person}{Joel~Z. Leibo}, {and} \bibinfo{person}{Audrunas Gruslys}.}
  \bibinfo{year}{2018}\natexlab{}.
\newblock \showarticletitle{{Deep Q-learning From Demonstrations}}. In
  \bibinfo{booktitle}{{\em Proceedings of the 32nd AAAI Conference on
  Artificial Intelligence}}.
\newblock


\bibitem[\protect\citeauthoryear{Konidaris, Kuindersma, Grupen, and
  Barto}{Konidaris et~al\mbox{.}}{2012}]%
        {Konidaris2011}
\bibfield{author}{\bibinfo{person}{George Konidaris}, \bibinfo{person}{Scott
  Kuindersma}, \bibinfo{person}{Roderic Grupen}, {and} \bibinfo{person}{Andrew
  Barto}.} \bibinfo{year}{2012}\natexlab{}.
\newblock \showarticletitle{{Robot Learning from Demonstration by Constructing
  Skill Trees}}.
\newblock \bibinfo{journal}{{\em International Journal of Robotics Research\/}}
  \bibinfo{volume}{31}, \bibinfo{number}{3} (\bibinfo{year}{2012}),
  \bibinfo{pages}{360--375}.
\newblock


\bibitem[\protect\citeauthoryear{Kulkarni, Narasimhan, Saeedi, and
  Tenenbaum}{Kulkarni et~al\mbox{.}}{2016}]%
        {Kulkarni2016}
\bibfield{author}{\bibinfo{person}{Tejas Kulkarni}, \bibinfo{person}{Karthik
  Narasimhan}, \bibinfo{person}{Ardavan Saeedi}, {and} \bibinfo{person}{Josh
  Tenenbaum}.} \bibinfo{year}{2016}\natexlab{}.
\newblock \showarticletitle{{Hierarchical Deep Reinforcement Learning:
  Integrating Temporal Abstraction and Intrinsic Motivation}}. In
  \bibinfo{booktitle}{{\em Advances in Neural Information Processing Systems
  29}}.
\newblock


\bibitem[\protect\citeauthoryear{Lake, Ullman, Tenenbaum, and Gershman}{Lake
  et~al\mbox{.}}{2017}]%
        {Lake2017}
\bibfield{author}{\bibinfo{person}{Brenden Lake}, \bibinfo{person}{Tomer
  Ullman}, \bibinfo{person}{Joshua Tenenbaum}, {and} \bibinfo{person}{Samuel
  Gershman}.} \bibinfo{year}{2017}\natexlab{}.
\newblock \showarticletitle{{Building Machines that Learn and Think Like
  People}}.
\newblock \bibinfo{journal}{{\em Behavioral and Brain Sciences\/}}
  \bibinfo{volume}{40} (\bibinfo{year}{2017}).
\newblock


\bibitem[\protect\citeauthoryear{Lyu, Yang, Liu, and Gustafson}{Lyu
  et~al\mbox{.}}{2019}]%
        {Lyu2019}
\bibfield{author}{\bibinfo{person}{Daoming Lyu}, \bibinfo{person}{Fangkai
  Yang}, \bibinfo{person}{Bo Liu}, {and} \bibinfo{person}{Steven Gustafson}.}
  \bibinfo{year}{2019}\natexlab{}.
\newblock \showarticletitle{{SDRL: Interpretable and Data-Efficient Deep
  Reinforcement Learning Leveraging Symbolic Planning}}. In
  \bibinfo{booktitle}{{\em Proceedings of the 33rd AAAI Conference on
  Artificial Intelligence}}.
\newblock


\bibitem[\protect\citeauthoryear{Machado, Bellemare, Talvitie, Veness,
  Hausknecht, and Bowling}{Machado et~al\mbox{.}}{2018}]%
        {Machado2018}
\bibfield{author}{\bibinfo{person}{Marlos Machado}, \bibinfo{person}{Marc
  Bellemare}, \bibinfo{person}{Erik Talvitie}, \bibinfo{person}{Joel Veness},
  \bibinfo{person}{Matthew Hausknecht}, {and} \bibinfo{person}{Michael
  Bowling}.} \bibinfo{year}{2018}\natexlab{}.
\newblock \showarticletitle{{Revisiting the Arcade Learning Environment}}. In
  \bibinfo{booktitle}{{\em Proceedings of the 27th International Joint
  Conference on Artificial Intelligence}}.
\newblock


\bibitem[\protect\citeauthoryear{Mnih, Kavukcuoglu, Silver, Rusu, Veness,
  Bellemare, Graves, Riedmiller, Fidjeland, Ostrovski, Petersen, Beattie,
  Sadik, Antonoglou, King, Kumaran, Wierstra, Legg, and Hassabis}{Mnih
  et~al\mbox{.}}{2015}]%
        {Mnih2015}
\bibfield{author}{\bibinfo{person}{Volodymyr Mnih}, \bibinfo{person}{Koray
  Kavukcuoglu}, \bibinfo{person}{David Silver}, \bibinfo{person}{Andrei Rusu},
  \bibinfo{person}{Joel Veness}, \bibinfo{person}{Marc Bellemare},
  \bibinfo{person}{Alex Graves}, \bibinfo{person}{Martin Riedmiller},
  \bibinfo{person}{Andreas Fidjeland}, \bibinfo{person}{Georg Ostrovski},
  \bibinfo{person}{Stig Petersen}, \bibinfo{person}{Charles Beattie},
  \bibinfo{person}{Amir Sadik}, \bibinfo{person}{Ioannis Antonoglou},
  \bibinfo{person}{Helen King}, \bibinfo{person}{Dharshan Kumaran},
  \bibinfo{person}{Daan Wierstra}, \bibinfo{person}{Shane Legg}, {and}
  \bibinfo{person}{Demis Hassabis}.} \bibinfo{year}{2015}\natexlab{}.
\newblock \showarticletitle{{Human-level Control through Deep Reinforcement
  Learning}}.
\newblock \bibinfo{journal}{{\em Nature\/}} \bibinfo{volume}{518},
  \bibinfo{number}{7540} (\bibinfo{year}{2015}), \bibinfo{pages}{529--533}.
\newblock


\bibitem[\protect\citeauthoryear{Ng, Harada, and Russell}{Ng
  et~al\mbox{.}}{1999}]%
        {Ng1999}
\bibfield{author}{\bibinfo{person}{Andrew Ng}, \bibinfo{person}{Daishi Harada},
  {and} \bibinfo{person}{Stuart Russell}.} \bibinfo{year}{1999}\natexlab{}.
\newblock \showarticletitle{{Policy Invariance Under Reward Transformations:
  Theory and Application to Reward Shaping}}. In \bibinfo{booktitle}{{\em
  Proceedings of the 16th International Conference on Machine Learning}}.
\newblock


\bibitem[\protect\citeauthoryear{Roderick, Grimm, and Tellex}{Roderick
  et~al\mbox{.}}{2018}]%
        {Roderick2018}
\bibfield{author}{\bibinfo{person}{Melrose Roderick},
  \bibinfo{person}{Christopher Grimm}, {and} \bibinfo{person}{Stefanie
  Tellex}.} \bibinfo{year}{2018}\natexlab{}.
\newblock \showarticletitle{{Deep Abstract Q-Networks}}. In
  \bibinfo{booktitle}{{\em Proceedings of the 17th International Conference on
  Autonomous Agents and MultiAgent Systems}}.
\newblock


\bibitem[\protect\citeauthoryear{Salimans and Chen}{Salimans and Chen}{2018}]%
        {Salimans2018}
\bibfield{author}{\bibinfo{person}{Tim Salimans} {and} \bibinfo{person}{Richard
  Chen}.} \bibinfo{year}{2018}\natexlab{}.
\newblock \showarticletitle{{Learning Montezuma's Revenge from a Single
  Demonstration}}. In \bibinfo{booktitle}{{\em Proceedings of the NeurIPS Deep
  Reinforcement Learning Workshop}}.
\newblock


\bibitem[\protect\citeauthoryear{Sammut, Hurst, Kedzier, and Michie}{Sammut
  et~al\mbox{.}}{1992}]%
        {Sammut1992}
\bibfield{author}{\bibinfo{person}{Claude Sammut}, \bibinfo{person}{Scott
  Hurst}, \bibinfo{person}{Dana Kedzier}, {and} \bibinfo{person}{Donald
  Michie}.} \bibinfo{year}{1992}\natexlab{}.
\newblock \showarticletitle{{Learning to Fly}}. In \bibinfo{booktitle}{{\em
  Proceedings of the 9th International Workshop on Machine Learning}}.
\newblock


\bibitem[\protect\citeauthoryear{Schaal}{Schaal}{1997}]%
        {Schaal1997}
\bibfield{author}{\bibinfo{person}{Stefan Schaal}.}
  \bibinfo{year}{1997}\natexlab{}.
\newblock \showarticletitle{{Learning from Demonstration}}. In
  \bibinfo{booktitle}{{\em Advances in Neural Information Processing Systems
  9}}.
\newblock


\bibitem[\protect\citeauthoryear{Sutton and Barto}{Sutton and Barto}{2018}]%
        {Sutton2018}
\bibfield{author}{\bibinfo{person}{Richard Sutton} {and}
  \bibinfo{person}{Andrew Barto}.} \bibinfo{year}{2018}\natexlab{}.
\newblock \bibinfo{booktitle}{{\em {Reinforcement Learning: An Introduction}\/}
  (\bibinfo{edition}{2} ed.)}.
\newblock \bibinfo{publisher}{MIT Press}, \bibinfo{address}{Cambridge, MA}.
\newblock


\bibitem[\protect\citeauthoryear{Sutton, Precup, and Singh}{Sutton
  et~al\mbox{.}}{1999}]%
        {Sutton1999}
\bibfield{author}{\bibinfo{person}{Richard Sutton}, \bibinfo{person}{Doina
  Precup}, {and} \bibinfo{person}{Satinder Singh}.}
  \bibinfo{year}{1999}\natexlab{}.
\newblock \showarticletitle{{Between MDPs and Semi-MDPs: A Framework for
  Temporal Abstraction in Reinforcement Learning}}.
\newblock \bibinfo{journal}{{\em Artificial Intelligence\/}}
  \bibinfo{volume}{112}, \bibinfo{number}{1} (\bibinfo{year}{1999}),
  \bibinfo{pages}{181 -- 211}.
\newblock


\bibitem[\protect\citeauthoryear{Taylor, Suay, and Chernova}{Taylor
  et~al\mbox{.}}{2011}]%
        {Taylor2011}
\bibfield{author}{\bibinfo{person}{Matthew Taylor},
  \bibinfo{person}{Halit~Bener Suay}, {and} \bibinfo{person}{Sonia Chernova}.}
  \bibinfo{year}{2011}\natexlab{}.
\newblock \showarticletitle{{Integrating Reinforcement Learning with Human
  Demonstrations of Varying Ability}}. In \bibinfo{booktitle}{{\em Proceedings
  of the 10th International Conference on Autonomous Agents and Multi-agent
  Systems}}.
\newblock


\bibitem[\protect\citeauthoryear{Torrey and Taylor}{Torrey and Taylor}{2013}]%
        {Torrey2013}
\bibfield{author}{\bibinfo{person}{Lisa Torrey} {and} \bibinfo{person}{Matthew
  Taylor}.} \bibinfo{year}{2013}\natexlab{}.
\newblock \showarticletitle{{Teaching on a Budget: Agents Advising Agents in
  Reinforcement Learning}}. In \bibinfo{booktitle}{{\em Proceedings of the 12th
  International Conference on Autonomous Agents and Multiagent Systems}}.
\newblock


\bibitem[\protect\citeauthoryear{van Seijen, Fatemi, Romoff, Laroche, Barnes,
  and Tsang}{van Seijen et~al\mbox{.}}{2017}]%
        {vanSeijen2017}
\bibfield{author}{\bibinfo{person}{Harm van Seijen}, \bibinfo{person}{Mehdi
  Fatemi}, \bibinfo{person}{Joshua Romoff}, \bibinfo{person}{Romain Laroche},
  \bibinfo{person}{Tavian Barnes}, {and} \bibinfo{person}{Jeffrey Tsang}.}
  \bibinfo{year}{2017}\natexlab{}.
\newblock \showarticletitle{{Hybrid Reward Architecture for Reinforcement
  Learning}}. In \bibinfo{booktitle}{{\em Advances in Neural Information
  Processing Systems 30}}.
\newblock


\bibitem[\protect\citeauthoryear{Vezhnevets, Osindero, Schaul, Heess,
  Jaderberg, Silver, and Kavukcuoglu}{Vezhnevets et~al\mbox{.}}{2016}]%
        {Vezhnevets2017}
\bibfield{author}{\bibinfo{person}{Alexander Vezhnevets},
  \bibinfo{person}{Simon Osindero}, \bibinfo{person}{Tom Schaul},
  \bibinfo{person}{Nicolas Heess}, \bibinfo{person}{Max Jaderberg},
  \bibinfo{person}{David Silver}, {and} \bibinfo{person}{Koray Kavukcuoglu}.}
  \bibinfo{year}{2016}\natexlab{}.
\newblock \showarticletitle{{FeUdal Networks for Hierarchical Reinforcement
  Learning}}. In \bibinfo{booktitle}{{\em Proceedings of the 34th International
  Conference on Machine Learning}}.
\newblock


\bibitem[\protect\citeauthoryear{Watkins and Dayan}{Watkins and Dayan}{1992}]%
        {Watkins1992}
\bibfield{author}{\bibinfo{person}{Christopher Watkins} {and}
  \bibinfo{person}{Peter Dayan}.} \bibinfo{year}{1992}\natexlab{}.
\newblock \showarticletitle{{Q-learning}}.
\newblock \bibinfo{journal}{{\em Machine Learning\/}} \bibinfo{volume}{8},
  \bibinfo{number}{3} (\bibinfo{year}{1992}), \bibinfo{pages}{279--292}.
\newblock


\bibitem[\protect\citeauthoryear{Xue, Lu, Levin, Weller, Li, and Bechara}{Xue
  et~al\mbox{.}}{2009}]%
        {Xue2009}
\bibfield{author}{\bibinfo{person}{Gui Xue}, \bibinfo{person}{Zhonglin Lu},
  \bibinfo{person}{Irwin Levin}, \bibinfo{person}{Joshua Weller},
  \bibinfo{person}{Xiangrui Li}, {and} \bibinfo{person}{Antoine Bechara}.}
  \bibinfo{year}{2009}\natexlab{}.
\newblock \showarticletitle{{Functional Dissociations of Risk and Reward
  Processing in the Medial Prefrontal Cortex}}.
\newblock \bibinfo{journal}{{\em Cerebral Cortex\/}} \bibinfo{volume}{19},
  \bibinfo{number}{5} (\bibinfo{year}{2009}), \bibinfo{pages}{1019–1027}.
\newblock


\bibitem[\protect\citeauthoryear{Zang, Zhou, Minnen, and Isbell}{Zang
  et~al\mbox{.}}{2009}]%
        {Zang2009}
\bibfield{author}{\bibinfo{person}{Peng Zang}, \bibinfo{person}{Peng Zhou},
  \bibinfo{person}{David Minnen}, {and} \bibinfo{person}{Charles Isbell}.}
  \bibinfo{year}{2009}\natexlab{}.
\newblock \showarticletitle{{Discovering Options from Example Trajectories}}.
  In \bibinfo{booktitle}{{\em Proceedings of the 26th International Conference
  on Machine Learning}}.
\newblock


\end{thebibliography}

\end{document}